\documentclass[journal,twoside,web]{ieeecolor}
\usepackage{tmi}
\usepackage{cite}
\usepackage{amsmath,amssymb,amsfonts}
\usepackage{algorithmic}
\usepackage{graphicx}
\usepackage{textcomp}
\usepackage{booktabs}
\usepackage{multirow}
\usepackage{color}
\usepackage{threeparttable}
\usepackage{makecell}
\usepackage{tabularx}
\usepackage[colorlinks,
            linkcolor=red,
            anchorcolor=blue,
            citecolor=green
            ]{hyperref}
\usepackage{float}
\def\BibTeX{{\rm B\kern-.05em{\sc i\kern-.025em b}\kern-.08em
    T\kern-.1667em\lower.7ex\hbox{E}\kern-.125emX}}
\graphicspath{ {./fig/} }
\begin{document}
\title{Large Language Model with Region-guided Referring and Grounding for CT Report Generation}
\author{Zhixuan Chen, Yequan Bie, Haibo Jin, and Hao Chen, \IEEEmembership{Senior Member, IEEE}
\thanks{This work was supported by the Hong Kong Innovation and Technology Fund (Project No. MHP/002/22), HKUST (Project No. FS111), and the Research Grants Council of the Hong Kong Special Administrative Region, China (Project Reference Number: T45-401/22-N).}
\thanks{Zhixuan Chen, Yequan Bie, and Haibo Jin are with the Department of Computer
Science and Engineering at the Hong Kong University of Science and Technology University, Hong Kong SAR, China. (e-mail: \{zchenhi, ybie\}@connect.ust.hk, hjinag@cse.ust.hk).}
\thanks{Hao Chen is with the Department of Computer Science and Engineering, Department of Chemical and Biological Engineering and Division of Life Science, Hong Kong University of Science and Technology, Hong Kong, China (e-mail: jhc@cse.ust.hk).}
\thanks{The corresponding author is Hao Chen.}
}

\maketitle

\begin{abstract}
Computed tomography (CT) report generation is crucial to assist radiologists in interpreting CT volumes, which can be time-consuming and labor-intensive. Existing methods primarily only consider the global features of the entire volume, making it struggle to focus on specific regions and potentially missing abnormalities. To address this issue, we propose Reg2RG, the first region-guided referring and grounding framework for CT report generation, which enhances diagnostic performance by focusing on anatomical regions within the volume. Specifically, we utilize masks from a universal segmentation module to capture local features for each referring region. A local feature decoupling (LFD) strategy is proposed to preserve the local high-resolution details with little computational overhead. Then the local features are integrated with global features to capture inter-regional relationships within a cohesive context. Moreover, we propose a novel region-report alignment (RRA) training strategy. It leverages the recognition of referring regions to guide the generation of region-specific reports, enhancing the model's referring and grounding capabilities while also improving the report's interpretability. A large language model (LLM) is further employed as the language decoder to generate reports from integrated visual features, facilitating region-level comprehension. Extensive experiments on two large-scale chest CT-report datasets demonstrate the superiority of our method, which outperforms several state-of-the-art methods in terms of both natural language generation and clinical efficacy metrics while preserving promising interpretability. The code is available at \href{https://github.com/zhi-xuan-chen/Reg2RG}{https://github.com/zhi-xuan-chen/Reg2RG}.
\end{abstract}


\section{Introduction}
\label{sec:introduction}

Computed tomography (CT) is widely used in clinical practice and crucial for diagnoses~\cite{hussain2022modern}. However, this process is labor-intensive~\cite{goergen2013evidence} as radiologists need to analyze entire CT volumes and produce detailed diagnostic reports. To alleviate the workload, automatic CT report generation has attracted increasing attention~\cite{hamamci2024ct2rep,wu2023towards,bai2024m3d}. 

\begin{figure}
    \centering
    \includegraphics[width=\linewidth]{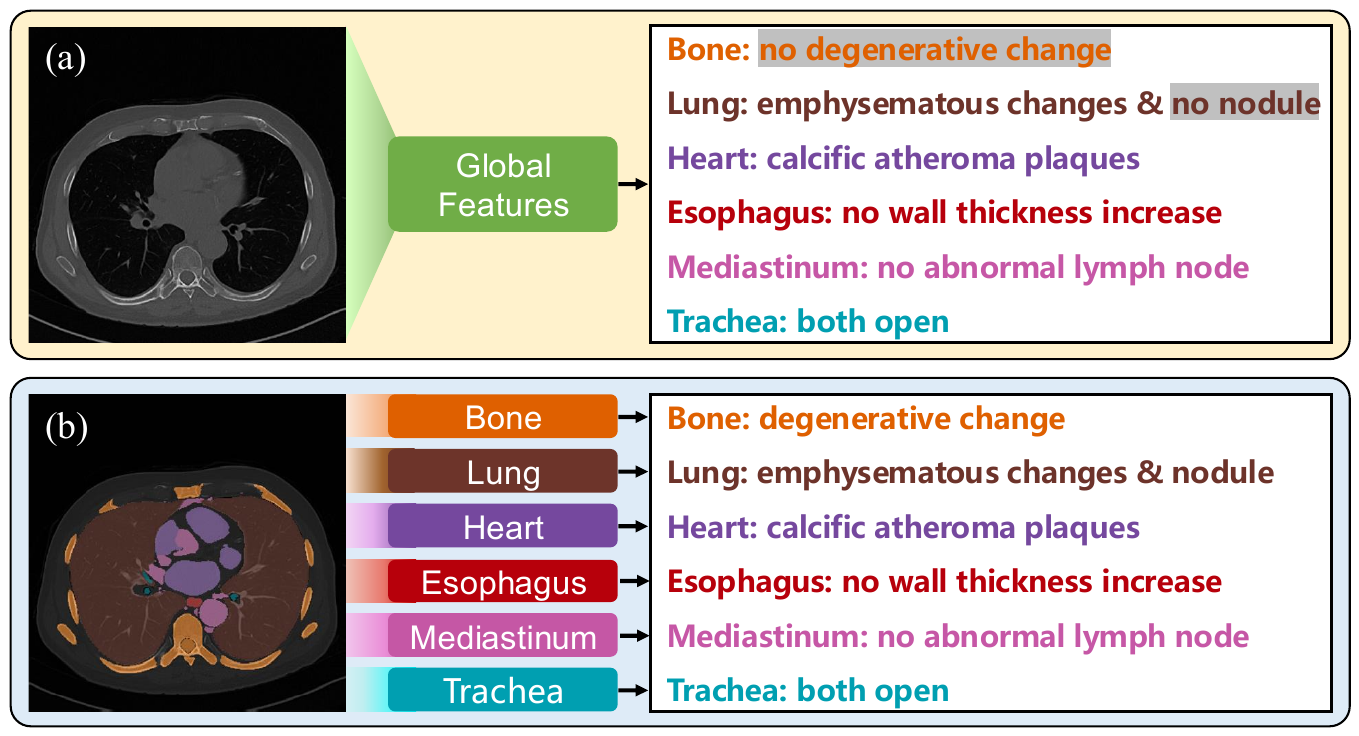}
    \caption{Our method vs. the vanilla method. The gray background highlights instances of incorrect diagnosis. For ease of comparison, we divide the report into region-level sections. (a) The vanilla method based on global features is prone to neglecting some abnormalities since it fails to explore local details. (b) In contrast, our method can correctly detect all abnormalities with the region-guided local features.}
    \label{fig:toy}
\end{figure}

CT report generation involves creating detailed and accurate diagnostic reports from CT volumes, where each anatomical region must be thoroughly analyzed and described to ensure reliable clinical insights. Existing methods~\cite{hamamci2024ct2rep,bai2024m3d,wu2023towards} primarily rely on global features extracted from the entire volume to generate reports. While effective to some extent, this approach overlooks the inherent complexity of CT data as a three-dimensional (3D) imaging modality. CT volumes encode rich anatomical details across various regions, presenting significant challenges for decoders to accurately capture region-specific abnormalities using only volume-level embeddings. As depicted in Fig.~\ref{fig:toy}~(a), the vanilla method often fails to identify certain abnormalities when limited to global features alone. Therefore, enhancing the model’s ability to process and integrate region-specific information is crucial for generating comprehensive and clinically relevant CT reports.

Recently, universal segmentation models~\cite{zhao2023one,ma2024segment,stock2024segment} have achieved remarkable progress and possess potent zero-shot capabilities. This inspires us to leverage these models as off-the-shelf tools to extract anatomical masks from CT volumes, providing key regional information for report generation. In this work, we propose \textbf{Reg2RG}, a \textbf{Re}gion-\textbf{g}uided \textbf{R}eferring and \textbf{G}rounding framework for \textbf{R}eport \textbf{G}eneration, leveraging the in-context learning and long-term referencing capabilities of LLMs. Referring and grounding are closely related but distinct concepts in visual understanding tasks. Referring focuses on understanding the semantics of designated regions within an image and providing their descriptions. In contrast, grounding involves locating specific regions based on textual information, effectively linking language to visual elements. With these abilities, the model can accurately refer to specific regions for detailed diagnoses and ground reports to the correct regions for better interpretability. As shown in Fig.~\ref{fig:toy}~(b), our method can generate more accurate reports, which are well-grounded in each referring region of the volume.


To reduce computational costs, previous works~\cite{bai2024m3d,wu2023towards} often downsample volumes or pool features. However, these methods may lose crucial high-resolution texture details~\cite{liu2024benchmarking}. To address this issue, we propose a local feature decoupling (LFD) strategy to preserve local high-resolution texture details and essential geometry with low computational cost. 

Since the global volume-level embedding provides significant inter-regional relationships within a cohesive context, we further integrate global features with local features to enable collaboration between them. However, achieving this effective global-local collaboration is non-trivial. Since global features encompass information from all regions, they may interfere with the diagnosis of a specific region, as validated in Table~\ref{tab:ablation}. To address this challenge, we propose a novel region-report alignment (RRA) training strategy that enhances the alignment between local features and their corresponding region reports, reducing the influence of irrelevant global information and ensuring that the generated reports are more robust and accurate.

In medicine, interpretability is vital for radiologists to comprehend the visual basis of the generated reports. Previous works~\cite{li2024organ,chen2020generating} mainly use attention maps to link report keywords with image regions. However, these attention maps are often coarse and ambiguous, lacking precision and reliability. The proposed RRA training strategy can also enhance interpretability by establishing a clear association between the referring region and the report. This explicit alignment ensures that the generated reports are firmly grounded in the identified regions, thereby improving both interpretability and reliability.


In summary, our contributions are as follows:

\begin{itemize}
    \item We propose \textbf{Reg2RG}, a novel \textbf{Re}gion-\textbf{g}uided \textbf{R}eferring and \textbf{G}rounding framework for \textbf{R}eport \textbf{G}eneration. It generates accurate reports by focusing on target regions from multiple candidates and aligning them with the correct regions for better interpretability. To our knowledge, this is the first work introducing referring and grounding for CT report generation.
    \item We design a local feature decoupling (LFD) strategy to decouple texture and geometry, preserving high-resolution texture details and essential geometry with minimal overhead. The global features are integrated with local features to collaboratively capture inter-regional relationships within a cohesive context.
    \item A region-report alignment (RRA) training strategy is utilized to boost the model’s referring and grounding abilities, making it more interpretable and reliable.
    \item We highlight the critical role of large-scale LLMs in report generation lies in their exceptional region-level referencing and grounding capabilities. These abilities allow for precise focus on region-specific features and inter-regional relationships, enabling the creation of more accurate and clinically relevant reports.
    \item Experiments on two large-scale 3D chest CT datasets demonstrate the effectiveness of our method, achieving both superior performance and interpretability.

\end{itemize}

\section{Related Works}
\label{sec:relate}
\subsection{Medical Report Generation}
To alleviate the heavy workload of pathologists, medical report generation has emerged as an effective solution for the automatic interpretation of medical images. The previous works~\cite{chen2020generating,chen2022cross,jin2024promptmrg,li2024organ} mostly focus on the 2D chest X-ray (CXR) report generation. To produce higher-quality reports, recent efforts~\cite{jin2024promptmrg,li2024organ,tanida2023interactive} have focused on incorporating additional information to enhance the accuracy of key abnormality details in the generated reports. For instance, PromptMRG~\cite{jin2024promptmrg} utilizes abnormality classification results as diagnostic prompts for the decoder, enhancing both the clinical relevance and effectiveness of the generated reports. Similarly, ORGAN~\cite{li2024organ} constructs an observation graph to better aggregate clinically significant information, further improving the quality and coherence of the reports. Building on this trend, RGRG~\cite{tanida2023interactive} utilizes a detector to introduce regional information, supporting fine-grained report generation. However, this approach has not been thoroughly explored in the more expansive and information-rich 3D space of CT volumes, where capturing regional details is particularly essential.

Leveraging large-scale CT-report datasets~\cite{hamamci2024foundation,zhang2024radgenome,tang2024work}, automatic CT report generation has also garnered increasing attention recently. CT2Rep~\cite{hamamci2024ct2rep} represents the first exploration into 3D CT report generation, leveraging a memory-driven decoder to generate detailed reports directly from global volume features. Inspired by the advancements in multi-modal LLMs~\cite{li2022blip,liu2024visual}, recent approaches have attempted to bootstrap LLMs for CT report generation. Dia-LLaMA~\cite{chen2024dia} integrates critical diagnostic prompts as prior medical knowledge, while HILT~\cite{liu2024benchmarking} emphasizes efficient encoding strategies for high-resolution volume features to enhance performance. Additionally, several studies~\cite{bai2024m3d,li2024towards,wu2023towards,zhang2023pmc} have explored the development of medical generalist models based on LLMs, demonstrating their capacity to perform CT report generation as part of broader diagnostic tasks.

However, these approaches generate reports solely based on global features, overlooking the critical role of local anatomical information. Furthermore, LLMs’ referencing and context-learning capabilities are underutilized. To address these issues, we propose a region-guided referring and grounding framework that leverages local features to enhance regional comprehension, fully harnessing the capabilities of LLM to optimize CT report generation.

\subsection{Region-level Referring and Grounding}

To promote region-level understanding, several works in the general domain~\cite{wang2024visionllm,zhang2023gpt4roi,ma2024groma,guo2024regiongpt,you2023ferret} have explored integrating LLMs with region-level features, showcasing their potential to enhance fine-grained reasoning and context-aware interpretations. However, the regional interpretation of medical images remains relatively underexplored. RGRG~\cite{tanida2023interactive} extracts region features for CXR report generation but generates reports independently for each region, neglecting inter-regional relationships and becoming inefficient with more regions. Some methods~\cite{bannur2024maira,alkhaldi2024minigpt,huang2024refer} attempt to ground diagnostic text to targeted regions in medical images, yet they fail to utilize region-level features for generating fine-grained, context-aware descriptions. MedRegA~\cite{wang2024interpretable} refers to specific regions using bounding box coordinates in prompts but may be less accurate compared to approaches that leverage detailed regional features. The ideal integration and utilization of regional information in medical imaging tasks remains an open question.

Given the importance of geometric information in assessing lesion size and position in medicine, we propose a novel local feature decoupling strategy that preserves complete geometric information while retaining detailed texture features. Inspired by Groma~\cite{ma2024groma}, we achieve grounded report generation by referencing regions from a pool of candidates, enabling focused and relevant analysis. Additionally, we integrate global features with local features to capture inter-regional relationships and provide a cohesive contextual understanding of the entire image. To enhance regional analysis, we propose a region-report alignment training strategy to explicitly link visual regions with their reports, enabling accurate and coherent multi-region report generation in a single inference.

\section{Methods}
\label{sec:methods}
In this section, we first overview our framework in Sec.~\ref{sec:overview}. Next, we detail the local feature decoupling (LFD) strategy in Sec.~\ref{sec:local}, and explain its integration with global features in Sec.~\ref{sec:collaboration}. Finally, we describe the region-report alignment (RRA) training strategy in Sec.~\ref{sec:strategy}.

\subsection{Overview}
\label{sec:overview}
\begin{figure*}[h]
    \centering
    \includegraphics[width=1\linewidth]{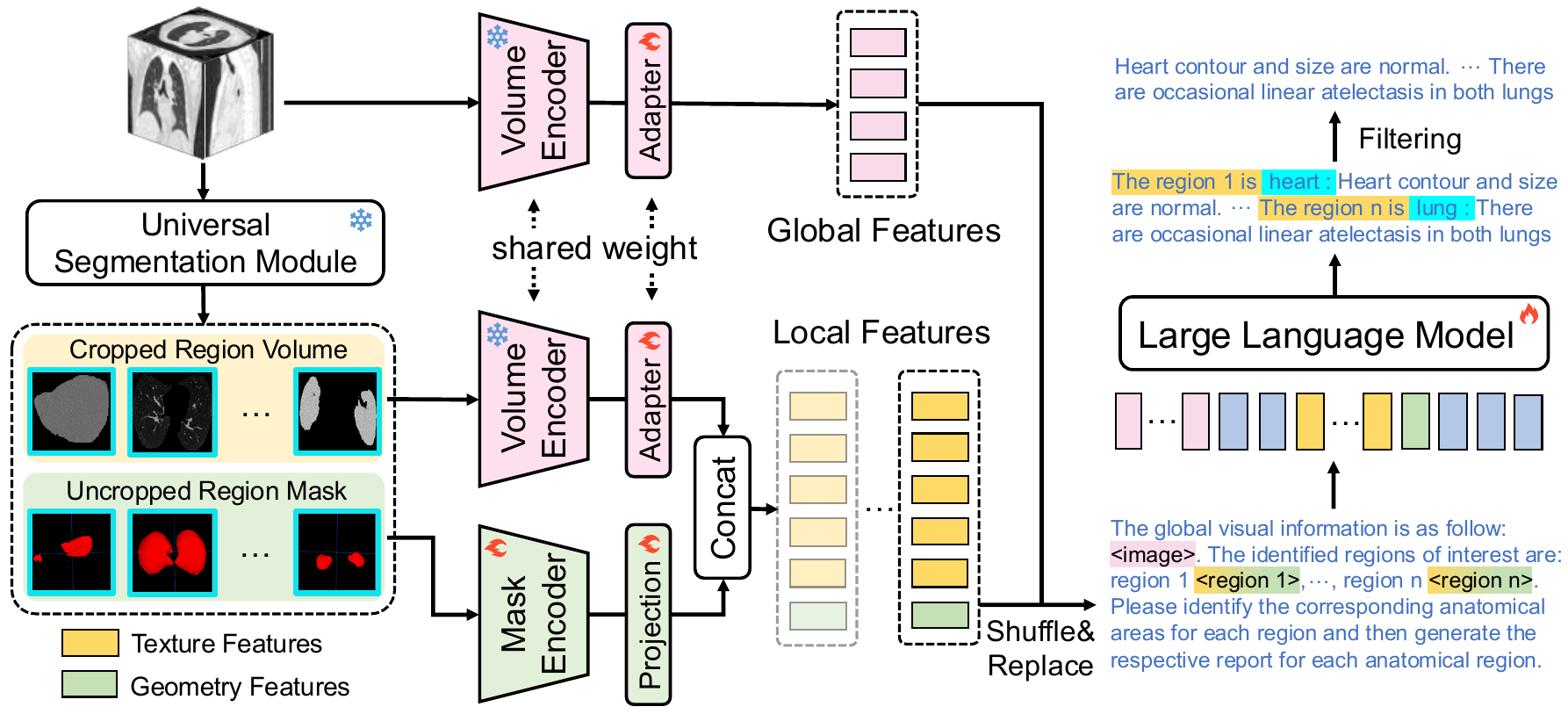}
    \caption{Overview of the proposed \textbf{Reg2RG} framework. It integrates global and local features as visual embeddings for the LLM to generate reports. Global features are encoded from the entire volume, while local features are extracted using segmentation masks to capture lesion details in sub-regions. The local features are decoupled into texture and geometry, where texture is derived from cropped masked volumes and geometry is obtained from the uncropped masks. Shuffling local features across various regions enhances the alignment between visual regions and their corresponding reports. The LLM focuses on each region individually to produce accurate and detailed region-specific reports.
    }
    \label{fig:framework}
\end{figure*}

The overview of our method \textbf{Reg2RG} is shown in Fig.~\ref{fig:framework}. CT report generation involves creating report $\mathbf{R}$ based on CT volumes $\mathbf{V}\in \mathbb{R}^{H\times W\times D}$. Unlike the previous works~\cite{hamamci2024ct2rep,bai2024m3d} that use only global features $\mathcal{G}$ of $\mathbf{V}$, we additionally extract a set of local features $\mathcal{L}=\{\mathcal{L}_1, \cdots, \mathcal{L}_n\}$ with the universal segmentation module. To preserve the local high-resolution texture details and significant geometry information with minimal computational overhead, local features are decoupled into texture and geometry. The decoupled local features work alongside global features to generate the report $\mathbf{R}$ using the LLM. Furthermore, we propose a training strategy that strengthens the alignment between the local feature $\mathcal{L}_i$ and the region-specific report $R_i$, thereby improving diagnostic accuracy and enhancing the reliability of the generated report. The overall generating process can be described as follows:
\begin{equation}
    \label{eq:1}
    \mathbf{R} =\text{LLM}(\mathcal{G}, \mathcal{L})=\text{LLM}(\mathcal{G}, \mathcal{L}_1,\cdots, \mathcal{L}_n),
\end{equation} 
where $n$ is the number of referring regions in $\mathbf{V}$.

\subsection{Local Feature Decoupling Strategy}
\label{sec:local}
As shown in Fig.~\ref{fig:framework}, we first utilize the existing universal segmentation module $f_S$ to extract mask $M_{A_j}$ for the anatomical area $A_j$ given the CT volumes $\mathbf{V}$. The segmentation process can be formulated as follows:
\begin{equation}
    \{M_{A_1},\cdots,M_{A_n}\} = f_S(\mathbf{V}).
\end{equation}

The region mask $M_{A_j}$ is then utilized to construct the corresponding local features $\mathcal{L}_{A_j}$. To preserve higher-resolution details without increasing much computational burden, we design a local feature decoupling (LFD) strategy that separates texture and geometry information.
Texture information refers to patterns within regions of interest that capture surface characteristics, which are essential for disease diagnosis. For example, in lung CT scans, the texture feature ``ground-glass opacity" indicates increased lung tissue density. Geometry information encompasses the size, shape, and spatial location of regions of interest, playing a crucial role in medical imaging. For example, the geometry feature ``enlarged heart" may indicate conditions such as heart failure or pericardial effusion.

For the texture information, we first use region mask $M_{A_j}$ to extract the region volume $V_{A_j}$ from $\mathbf{V}$ by element-wise multiplication. This results in a large redundant area outside the region of interest, which is not informative. Therefore, we crop the region volume $V_{A_j}$ to exclude these irrelevant parts. Since $V_{A_j}$ is much smaller than the entire volume $\mathbf{V}$, we can retain higher-resolution details without increasing the input size. Next, we utilize a 3D volume encoder $f_V$ to extract local texture features $\mathcal{L}_{A_j}^{t}$. The following adapter module $f_A$ is used to compress and align these local texture features with the embedding space of the LLM. This process can be formulated as follows:
\begin{equation}
    \mathcal{L}_{A_j}^{t} = f_A(f_V(\text{Crop}(V_{A_j})))= f_A(f_V(\text{Crop}(M_{A_j}\odot \mathbf{V}))),
\end{equation}
where $\odot$ indicates element-wise multiplication.

Previous works~\cite{ma2024groma,guo2024regiongpt,you2023ferret,wang2024visionllm,zhang2023gpt4roi} extract local features by only encoding cropped regions, focusing exclusively on texture features while neglecting the essential geometry features necessary for assessing lesion size and location in medical contexts. In contrast, our method incorporates them as supplementary features. Specifically, we introduce the geometry information by encoding the region mask $M_{A_j}$, which is uncropped to preserve the original size and position. A lightweight mask encoder $f_M$ is used to extract geometry features $\mathcal{L}_{A_j}^{g}$, followed by a fully connected layer $f_P$ to project these features for LLM input. This process is formulated as follows:
\begin{equation}
    \mathcal{L}_{A_j}^{g} = f_P(f_M(M_{A_j})).
\end{equation}

The local features $\mathcal{L}_{A_j}$ are obtained by concatenating the texture features $\mathcal{L}_{A_j}^{t}$ and geometry features $\mathcal{L}_{A_j}^{g}$:
\begin{equation}
    \mathcal{L}_{A_j} = \text{Concat}(\mathcal{L}_{A_j}^{t}, \mathcal{L}_{A_j}^{g}).
\end{equation}

Typically, the local features $\mathcal{L}_A$ comprise multiple regions $\mathcal{L}_{A_j}$, each providing specific information for a distinct anatomical area within the CT volumes:
\begin{equation}
    \mathcal{L}_A = \{\mathcal{L}_{A_1}, \cdots, \mathcal{L}_{A_n}\}.
\end{equation} 

Note that $\mathcal{L}_A$ is not identical to $\mathcal{L}$ in Eq.~(\ref{eq:1}); further details will be provided in Sec.~\ref{sec:strategy}.

\subsection{Global-Local Features Collaboration}
\label{sec:collaboration}
In medicine, different regions are interrelated rather than isolated. Therefore, we incorporate global features to provide contextual information. Specifically, the same volume encoder $f_V$ and adapter $f_A$ (as in Sec.~\ref{sec:local}) are used to extract global features $\mathcal{G}$:
\begin{equation}
    \mathcal{G} = f_A(f_V(\mathbf{V})).  
\end{equation}

The collaboration of global and local features is achieved by embedding them into the prompt. Our designed prompt $\mathcal{P}$ consists of two parts: $\mathcal{P}=\{\mathcal{I},\mathcal{T}\}$, where $\mathcal{I}$ denotes the special tokens for the visual embedding and $\mathcal{T}$ represents the text tokens of the instruction. As depicted in Fig.~\ref{fig:framework}, we utilize \textless image\textgreater\ and \textless region $i$\textgreater\ as the special tokens for the global and local features, respectively. These special tokens are replaced by the corresponding features $\mathcal{G}$ and $\mathcal{L}_i$, which then interact within the LLM to generate $\mathbf{R}$.

\subsection{Region-Report Alignment Training Strategy}
\label{sec:strategy}
To enhance the explicit link between the referring region and the report, we propose a training strategy that first recognizes the anatomical area of the referring region and then generates the corresponding region report $R_i$.

While the segmentation module can provide the anatomical area name, this information may introduce bias during report generation, as the model might rely on the name rather than the actual local features. For instance, if the input includes the name ``lung" for a region, the model may focus solely on generating content related to the ``lung", potentially overlooking the real information within the local features. Therefore, we train the model to recognize this information independently from the local features. This strategy helps the model better understand the referred region, making the generated report more reliably grounded. Furthermore, we shuffle the order of the local features $\mathcal{L}_{A}$ at each step to prevent the model from associating anatomical areas with a fixed sequence. Thus, although the content of \(\mathcal{L}\) and \(\mathcal{L}_A\) remains unchanged, the order of each local feature varies between them. The generation process is defined as follows:
\begin{align}
    \mathbf{R} &= \{(P_1, R_1), \cdots, (P_n, R_n)\} \notag\\
    &= \text{LLM}(\mathcal{G}, \mathcal{L}_{1}, \cdots, \mathcal{L}_{n}) \\
    &= \text{LLM}(\mathcal{G}, \text{Shuffle}(\mathcal{L}_{A_1}, \cdots, \mathcal{L}_{A_n})). \notag
\end{align} 

During training, region-level reports $R_i$ of each referring region are used as ground truth. In addition, we add a prefix $P_i$=``\emph{The region} [$i$]\ \emph{is}\ [\emph{area name}]'' before $R_i$ to indicate the anatomical area name. The model learns to recognize the anatomical area name by predicting this prefix. Since anatomical region names are fixed for each specific region, they provide a clear and structured target for the model, enabling it to develop a more precise semantic understanding of the referred regions. The restructured report remains a sequence of text tokens: $\mathbf{R} = \{(P_1, R_1), \cdots, (P_n, R_n)\} = \{r_1, r_2, \cdots, r_T\}$ ($T$ is the report length). This format integrates seamlessly with the native auto-regressive training process of the LLM. The training process is optimized by minimizing the language-modeling loss as shown below:
\begin{equation}
    \mathcal{L}_{\text{LM}} = -\sum_{t=1}^{T} \log p(r_t|\mathcal{P}, r_1, \cdots, r_{t-1}),
\end{equation}
where $\mathcal{P}$ indicates the prompt, as mentioned in Sec.~\ref{sec:collaboration}.

When evaluating the quality of the generated report, the prefix \( P_i \) is removed from the $\mathbf{R}$.

\section{Experiments and Results}
\subsection{Datasets and Evaluation Metrics} 
\label{sec:datasets}
\textbf{Datasets.}
We train and evaluate our model alongside comparative methods using two large-scale chest CT datasets, ensuring a more comprehensive assessment.

The RadGenome-ChestCT dataset~\cite{zhang2024radgenome}, designed for region-guided 3D chest CT interpretation, is based on the CT-RATE~\cite{hamamci2024foundation} dataset. It consists of 25,692 region-guided CT-report pairs sourced from 21,304 patients. The CT volumes possess a consistent voxel spacing of 1 mm $\times$ 1 mm $\times$ 3 mm, with anatomical masks generated using the SAT~\cite{zhao2023one} segmentation module. Region-grounded reports are generated with GPT-4~\cite{achiam2023gpt} and a named entity recognition model, covering 10 chest anatomical regions: abdomen, bones, breasts, esophagus, heart, lungs, trachea and bronchi, mediastinum, pleura, and thyroid. Following the official data split, 24,128 pairs are allocated for training, while 1,564 pairs are reserved for evaluation.

The CTRG-Chest-548K dataset~\cite{tang2024work} includes 1,804 CT volume-report pairs. To extract anatomical masks, we employ the segmentation module SAT~\cite{zhao2023one}. Since region-level grounded reports are not provided within this dataset, we utilize Qwen2.5-14B~\cite{qwen2} to segment the reports into region-level sections. Regarding the data split, the dataset is randomly partitioned into training and testing sets in an 8:2 proportion.

\textbf{Evaluation Metrics.}
In line with prior studies~\cite{chen2024dia,liu2024benchmarking,hamamci2024ct2rep}, we employ widely recognized natural language generation (NLG) metrics like BLEU-n~\cite{papineni2002bleu}, METEOR~\cite{banerjee2005meteor}, and ROUGE-L~\cite{lin2004rouge} for evaluation. BLEU-n measures the overlap of n-grams (sequences of n words) between generated and reference reports to gauge word sequence similarity. ROUGE-L assesses by comparing the longest common word subsequence, focusing on textual alignment. METEOR effectively incorporates synonyms and paraphrases, offering a more nuanced assessment of semantic similarity between reports.

NLG metrics focus on word and sentence similarity but neglect the diagnostic accuracy. For example, ``The heart is enlarged'' and ``The heart is not enlarged'' can have similar NLG scores despite opposite diagnostic conclusions. Therefore, we also adopt the clinical efficacy (CE) metrics~\cite{chen2020generating}. Following CT-CLIP~\cite{hamamci2024foundation}, we employ the RadBERT~\cite{yan2022radbert} text classifier to extract 18 types of abnormality labels from chest CT reports. As this classifier is fine-tuned on CT-RATE~\cite{hamamci2024foundation}, it ensures high-quality labels for the RadGenome-ChestCT dataset~\cite{zhang2024radgenome}, derived from CT-RATE. CE metrics are the precision, recall, and F1 score from comparing abnormality labels of generated and ground-truth reports, providing a clinically meaningful assessment by highlighting significant abnormalities. Since RadBERT is not specifically designed for the CTRG-Chest-548K dataset, the types of abnormalities RadBERT focuses on do not align with the primary abnormalities in this dataset. Therefore, we only use the CE metrics for the RadGenome-ChestCT dataset.

\subsection{Implementation Details and Baselines}
We use the text-prompted segmentation model SAT-Pro~\cite{zhao2023one} to extract anatomical masks when unavailable. The SAT-Pro model is configured with 256 object queries, and its input CT intensities are normalized to the range [0, 1] using min-max normalization. For our input, both global and local volumes are resized to 256 $\times$ 256 $\times$ 64. We use the pre-trained ViT3D and Perceiver from RadFM~\cite{wu2023towards} as the volume encoder $f_V$ and adapter $f_A$ to extract texture features, each represented by 32 visual embeddings. Given that geometric information is sparser than texture information, a lightweight 3-layer ViT3D serves as the mask encoder $f_M$ to extract geometric features. These features are pooled into a single embedding and projected by $f_P$ to align with the LLM's embedding space. For the LLM, we choose LLaMA2-7B~\cite{touvron2023llama} and utilize LoRA~\cite{hu2021lora} for parameter-efficient fine-tuning. The LoRA configuration is set with a rank \(r = 8\), a scaling factor \(\alpha = 32\), and a dropout rate of 0.1.

We train the model with the AdamW optimizer~\cite{loshchilov2017decoupled} at an initial learning rate of $5 \times 10^{-5}$, following a constant learning rate schedule with a warmup phase. Training on the RadGenome-Chest dataset takes 48 hours on two RTX 4090 GPUs with PyTorch 2.0, running for 6 epochs with an effective batch size of 16. For the CTRG-Chest-584K dataset, training takes 24 hours for 10 epochs with the same batch size. To optimize memory usage, ZeRO~\cite{rajbhandari2020zero} stage 2 is applied alongside gradient checkpointing~\cite{chen2016training}. We use the checkpoint of the last epoch for evaluation.

For baseline comparisons, we choose three state-of-the-art 3D methods capable of generating CT reports: CT2Rep~\cite{hamamci2024ct2rep}, RadFM~\cite{wu2023towards}, and M3D~\cite{bai2024m3d}. We also include two 2D methods R2GenGPT~\cite{wang2023r2gengpt} and MedVInT~\cite{zhang2023pmc}, which support radiology report generation. For fairness, we employ LLaMA2-7B~\cite{touvron2023llama} as the language decoder for all compared methods. Input volumes were resized to \(256 \times 256 \times 64\) and represented by 32 visual tokens for all methods. To adapt the 2D methods, we convert the 3D volumes into 2D multi-channel images. Each compared model is initialized with their pre-trained weights and is further fine-tuned on the CT report generation dataset.

\subsection{Quantitative Results}

\subsubsection{Natural Language Generation Metrics}
Table~\ref{tab:radgenome} presents the NLG results of our model alongside comparisons to other methods. On the RadGenome-ChestCT dataset, our model outperforms all others across all NLG metrics, underscoring its capability to generate high-quality reports. Specifically, we take the MedVInT~\cite{zhang2023pmc} with second-best results as an example. Our model achieves a relative improvement of 1.1\% to 6.7\% on BLEU metrics, highlighting enhanced expression similarity to the reference reports. For the METEOR metric considering inflectional variations and synonym matching, our model shows a notable 9.1\% improvement, indicating better lexical flexibility and semantic alignment. Our model also surpasses the second-best by 12.4\% in ROUGE-L, highlighting its consistent performance across metrics. 
A similar pattern is observed on the CTRG-Chest-584K dataset, where our method outperforms the second-best model with a 2.0\% to 3.7\% improvement in BLEU metrics and a 0.8\% gain in METEOR. The lower ROUGE-L score on this dataset may be attributed to the fragmented nature of region-level report generation. This fragmentation affects the coherence between reports for different regions and consequently impacts the evaluation of the longest common sequence. However, this inconsistency does not affect the quality of individual region reports, as evidenced by higher performance on other metrics. These results highlight the effectiveness of our model in generating high-quality reports.

\setlength{\tabcolsep}{2.5mm}  
\begin{table*}[t]  
\caption{The NLG performance of our model compared with other SOTA methods on two large-scale datasets. The best and second-best results are in \textbf{bold} and \underline{underlined}, respectively. A higher value indicates better performance.}
\centering  
\fontsize{10}{13}\selectfont  
\begin{threeparttable}  
	  
\begin{tabular}{cccp{1.1cm}<{\centering}p{1.1cm}<{\centering}p{1.1cm}<{\centering}p{1.1cm}<{\centering}p{1.1cm}<{\centering}p{1.3cm}<{\centering}}  

    \toprule\hline
    
    \bf Dataset & \bf Method & \bf Year & \bf BL-1  &\bf BL-2  &\bf BL-3   &\bf BL-4  &\bf MTR  &\bf RG-L  \cr

    \midrule 
    \multirow{6}{*}{RadGenome-ChestCT} 
     &  R2GenGPT~\cite{wang2023r2gengpt} &2023 & 43.28 & 34.11 & 28.16 & 24.16 & 39.85 & 32.26 \cr
    &  MedVInT~\cite{zhang2023pmc} &2023 & 44.28 & \underline{34.91} & \underline{28.75} & \underline{24.60} & \underline{40.39} & 32.58 \cr  
    & RadFM~\cite{wu2023towards} &2023 & 44.20  & 34.49 & 28.06 & 23.65 & 39.94 & 31.53  \cr   
    & CT2Rep~\cite{hamamci2024ct2rep} &2024 & \underline{44.42} & 34.43 & 27.94 & 23.56 & 40.16 & 30.99 \cr  
    & M3D~\cite{bai2024m3d} &2024 & 43.57 & 34.48 & 28.54 & 24.49 & 39.95 & \underline{32.61} \cr 
    \cmidrule(lr){2-9} 
    &  \textbf{Reg2RG} &- & \textbf{47.25} & \textbf{36.49} & \textbf{29.57} & \textbf{24.87} & \textbf{44.07} & \textbf{36.65} \cr  

    \midrule 
    \multirow{6}{*}{CTRG-Chest-584K} 
    &  R2GenGPT~\cite{wang2023r2gengpt} &2023 & 41.82 & 36.37 & 32.70 & 30.10 & 47.05 & \textbf{50.93} \cr
    &  MedVInT~\cite{zhang2023pmc} &2023 & 47.38 & 39.60 & 34.28 & 30.68 & \underline{49.32} & 49.53 \cr  
    & RadFM~\cite{wu2023towards}  &2023 & \underline{48.66}  & \underline{40.28} & \underline{34.73} & \underline{30.89} & 49.18 & 49.08  \cr   
    & CT2Rep~\cite{hamamci2024ct2rep}  &2024 & 42.28 & 36.16 & 32.08 & 29.19 & 47.00 & 50.17 \cr  
    & M3D~\cite{bai2024m3d} &2024 & 46.27 & 39.02 & 34.23 & 30.86 & 49.26 & \underline{50.24} \cr  
    \cmidrule(lr){2-9} 
    &  \textbf{Reg2RG} &- & \textbf{49.63} & \textbf{41.43} & \textbf{35.91} & \textbf{32.04} & \textbf{49.71} & 47.76 \cr  
    
    \hline\bottomrule
   
\end{tabular}  
\end{threeparttable} 
\label{tab:radgenome} 
\end{table*}

\subsubsection{Clinical Efficacy Metrics}
Table~\ref{tab:radgenome_ce} showcases the CE performance of our model compared to other methods on the RadGenome-ChestCT dataset. Our model surpasses the second-best approach by 3.9\%, 22.3\%, and 19.3\% in precision, recall, and F1 score, respectively. This demonstrates the superiority of our model in generating reports with higher diagnostic accuracy. It is worth noting that there is an inherent trade-off between precision and recall. Achieving higher precision requires minimizing false positives, which often leads the model to adopt a more conservative approach to predicting abnormalities. On the other hand, higher recall necessitates reducing false negatives, encouraging a more aggressive stance in abnormality detection. Therefore, balancing these competing priorities is particularly challenging, as demonstrated by the performance of other models. For instance, MedVInT~\cite{zhang2023pmc} achieves the second-highest recall but struggles with relatively low precision, while M3D~\cite{bai2024m3d} exhibits the reversed trend, favoring precision at the expense of recall. In contrast, our model effectively balances this trade-off, maintaining high performance in both metrics and achieving a significantly improved F1 score. These results underscore our model’s ability to maintain clinical relevance and diagnostic reliability while delivering high linguistic quality in the generated reports.

\setlength{\tabcolsep}{2.5mm}  
\begin{table}[t]  
\caption{The clinical efficacy performance of our model and other SOTA methods on the RadGenome-ChestCT~\cite{zhang2024radgenome} dataset. The best and second-best results are highlighted in \textbf{bold} and \underline{underlined}, respectively. A higher value implies better performance across all metrics.}
\centering  
\fontsize{10}{13}\selectfont  
\begin{threeparttable}  
	  
\begin{tabular}{cp{1.1cm}<{\centering}p{1.1cm}<{\centering}p{1.1cm}<{\centering}}  

    \toprule\hline
    
    \bf Method & \bf Pre.  &\bf Rec.  &\bf F1 \cr

    \midrule
    
    R2GenGPT~\cite{wang2023r2gengpt}  & 0.340 & 0.066 & 0.110  \cr
    
    MedVInT~\cite{zhang2023pmc} & 0.377 & \underline{0.148} & \underline{0.212}  \cr  
     RadFM~\cite{wu2023towards} & 0.382 & 0.131  &  0.195   \cr 
    CT2Rep~\cite{hamamci2024ct2rep}  & 0.317 & 0.089 & 0.139  \cr  
    M3D~\cite{bai2024m3d} & \underline{0.407} & 0.090 & 0.148  \cr   
    \midrule
        \textbf{Reg2RG} & \textbf{0.423} & \textbf{0.181} & \textbf{0.253}  \cr   
    \hline\bottomrule
   
\end{tabular}  
\end{threeparttable} 
\label{tab:radgenome_ce} 
\end{table}

\subsubsection{Region Recognition Results}
Table~\ref{tab:region_recognition} presents the region recognition performance of our model on the RadGenome-ChestCT dataset~\cite{zhang2024radgenome}. To enhance the model's interpretability and reliability of generated reports, our approach explicitly requires the model to first identify the anatomical area corresponding to the referring region before generating the associated report. This intermediate recognition step simplifies evaluation and interpretation compared to directly analyzing the generated reports, providing an early-stage validation of both their alignment with the target regions and the reliability of the reports. The results show that our model accurately identifies most anatomical regions except for the lung and pleura, suggesting the generated reports are reliably aligned with the target regions, thereby enhancing interpretability. The subpar results for the lung and pleura are attributed to the performance of the SAT~\cite{zhao2023one} segmentation module, which struggles to produce masks that adequately differentiate between these closely related regions. Once the segmentation model provides correct region masks, our model can effectively identify referring regions, leading to more trustworthy and clinically reliable reports.

\setlength{\tabcolsep}{2.5mm}  
\begin{table}[t]  
\caption{The region recognition results of our model. The Tra. \& Bro. region refers to the trachea and bronchi.}
\centering  
\fontsize{10}{13}\selectfont  
\begin{threeparttable}  
	  
\begin{tabular}{p{2.2cm}<{\centering}|p{1.3cm}<{\centering}p{1.3cm}<{\centering}p{1.3cm}<{\centering}}  

    \toprule\hline
    \bf Region&
    \bf Pre.&
    \bf Rec.&
    \bf F1\cr

    \midrule
    Abdomen & 0.997 & 0.996 & 0.997 \cr
    Bone & 0.996 & 0.998 & 0.997 \cr
    Breast & 0.945 & 0.983 & 0.964 \cr
    Esophagus & 0.997 & 0.999 & 0.998 \cr
    Heart & 0.995 & 0.996 & 0.996\cr
    Lung & 0.443 & 0.443 & 0.443\cr
    Mediastinum & 0.991 & 0.997 & 0.994 \cr
    Pleura & 0.442 & 0.441 & 0.442 \cr
    Thyroid & 0.991 & 0.931 & 0.960 \cr
    Tra. \& Bro. & 0.986 & 0.990 & 0.988 \cr
    
    \hline\bottomrule
   
\end{tabular}  
\end{threeparttable} 
\label{tab:region_recognition} 
\end{table}

\subsubsection{Region-level Reports Evaluation}
Table~\ref{tab:region_report} showcases the evaluation results of the region-level reports generated by our model on the RadGenome-ChestCT dataset~\cite{zhang2024radgenome}. We observe that the performance varies for reports from different anatomical regions. The higher occurrence frequency of certain regions in reports contributes to better performance on the abdomen, mediastinum, heart, and lung. In contrast, performance tends to be lower for regions that appear less frequently, such as the breast and thyroid. The subpar CE performance on the bone and trachea \& bronchi may be attributed to the RadBERT model from CT-CLIP~\cite{hamamci2024foundation} rarely considering abnormalities in these regions. Investigating approaches to improve the diagnosis of less common regions and abnormalities could serve as a direction for future research.

\setlength{\tabcolsep}{2.5mm}  
\begin{table*}[t]  
\caption{The evaluation of region-level reports generated by our model. The Tra. \& Bro. region refers to the trachea and bronchi. The \textbf{Amount} represents the number of reports that include descriptions for the specific region.}
\centering  
\fontsize{10}{13}\selectfont  
\begin{threeparttable}  
	  
\begin{tabular}{c|c|p{1.2cm}<{\centering}p{0.9cm}<{\centering}p{1.3cm}<{\centering}|p{1.25cm}<{\centering}p{1.25cm}<{\centering}p{1.3cm}<{\centering}p{1.65cm}<{\centering}}  

    \toprule\hline
    \multirow{2}{*}{\bf Region}&
    \multirow{2}{*}{\bf Amount}&
    \multicolumn{3}{c|}{\bf CE Metrics}&
    \multicolumn{4}{c}{\bf NLG Metrcis}\cr
    
    &&\bf Precision  &\bf Recall   &\bf F1-score  &\bf BLEU-1 &\bf BLEU-4  &\bf METEOR  &\bf ROUGE-L  \cr

    \midrule
    Abdomen & 1517 & 0.546 & 0.288 & 0.377 & 39.50 & 25.13 & 43.33 & 38.52 \cr
    Bone & 1509 & 0.000 & 0.000 & 0.000 & 41.05 & 28.49 & 47.15 & 42.13 \cr
    Breast & 56 & 0.174 & 0.211 & 0.190 & 22.46 & 13.63 & 23.14 & 21.96 \cr
    Esophagus & 1323 & 0.222 & 0.021 & 0.039 & 59.58 & 45.16 & 56.39 & 57.49 \cr
    Heart & 1418 & 0.320 & 0.135 & 0.190 & 39.85 & 27.44 & 44.75 & 43.19 \cr
    Lung & 1514 & 0.393 & 0.126 & 0.191 & 27.22 & 15.51 & 31.74 & 30.82 \cr
    Mediastinum & 1513 & 0.557 & 0.242 & 0.337 & 37.65 & 20.09 & 41.98 & 33.68 \cr
    Pleura & 1169 & 0.333 & 0.069 & 0.115 & 37.20 & 26.91 & 38.84 & 47.21 \cr
    Thyroid & 42 & 0.125 & 0.040 & 0.061 & 21.83 & 10.64 & 19.49 & 19.47 \cr
    Tra. \& Bro. & 1401 & 0.000 & 0.000 & 0.000 & 48.42 & 38.93 & 61.47 & 56.18 \cr

    \hline\bottomrule
   
\end{tabular}  
\end{threeparttable} 
\label{tab:region_report} 
\end{table*}

\subsection{Ablation Study}
To demonstrate the effectiveness of each component in our model, we conduct comprehensive ablation studies on the RadGenome-ChestCT dataset~\cite{zhang2024radgenome}. Given that our volume encoder $f_V$ and adapter $f_A$ are initialized using the pre-trained checkpoint from RadFM~\cite{wu2023towards}, we take it as the baseline for comparison. First, we examine the contribution of the local feature decoupling (LFD) strategy in Sec.~\ref{sec:ablation_local}. Next, we investigate the effectiveness of global-local features collaboration by incrementally adding different visual features in Sec.~\ref{sec:ablation_collaboration}. Additionally, we analyze the performance of our region-report alignment (RRA) training strategy in Sec.~\ref{sec:ablation_strategy}. Finally, we validate the necessity of employing large-scale LLMs for region-level referring and grounding in Sec.~\ref{sec:ablation_decoder}. The results of these experiments are detailed in Table~\ref{tab:decoupled}, Table~\ref{tab:ablation}, Table~\ref{tab:rra} and Table~\ref{tab:decoder}, demonstrating the individual and collective impact of each component.

\subsubsection{Local Feature Decoupling Strategy}
\label{sec:ablation_local}
First, we validate the efficacy of the local feature decoupling (LFD) strategy in improving model performance. In the baseline setup where decoupling is not applied, the model uses masked volumes without cropping as local features, combining texture and geometry information without separation. As Table~\ref{tab:decoupled} shows, the model with decoupled features demonstrates superior performance across most metrics. The significant improvement in CE metrics highlights the value of local high-resolution details from decoupled texture information in enhancing diagnostic accuracy. The slight decreases in BLEU-4 and ROUGE-L scores may be due to these metrics relying only on word overlap, which often fails to capture nuanced semantic similarities. By contrast, the higher METEOR score supports this assumption because it takes synonym matching and semantic relationships into consideration.

\setlength{\tabcolsep}{2.5mm}  
\begin{table*}[h]  
\caption{The effectiveness of local feature decoupling (LFD) strategy.}
\centering  
\fontsize{10}{13}\selectfont  
\begin{threeparttable}  
	  
\begin{tabular}{c|p{1.6cm}<{\centering}p{1.2cm}<{\centering}p{1.6cm}<{\centering}|p{1.6cm}<{\centering}p{1.6cm}<{\centering}p{1.8cm}<{\centering}p{2.1cm}<{\centering}}  

    \toprule\hline
    \multirow{2}{*}{\makecell{\bf LFD}}&
    \multicolumn{3}{c|}{\bf CE Metrics}&
    \multicolumn{4}{c}{\bf NLG Metrcis}\cr
    
    &\bf Precision  &\bf Recall   &\bf F1-score  &\bf BLEU-1 &\bf BLEU-4  &\bf METEOR  &\bf ROUGE-L  \cr

    \midrule
    & 0.349 & 0.140 & 0.200 & 46.85 & \textbf{25.69} & 43.74 & \textbf{37.64} \cr
    \checkmark & \textbf{0.423} & \textbf{0.181} & \textbf{0.253} & \textbf{47.25} & 24.87 & \textbf{44.07} & 36.65 \cr

    \hline\bottomrule
   
\end{tabular}  
\end{threeparttable} 
\label{tab:decoupled} 
\end{table*}

	  

    


   

\setlength{\tabcolsep}{2.6mm}  
\begin{table*}[t]  
\caption{The effectiveness of different visual features. The best results are highlighted in \textbf{bold}. The \textbf{TXT} and \textbf{GEO} denote the texture and geometric information of local features, respectively, whereas \textbf{GLB} stands for the global features. }
\centering  
\fontsize{10}{13}\selectfont  
\begin{threeparttable}  
	  
\begin{tabular}{c|p{0.8cm}<{\centering}p{0.8cm}<{\centering}p{0.8cm}<{\centering}|p{0.9cm}<{\centering}p{0.9cm}<{\centering}p{0.9cm}<{\centering}|p{1.cm}<{\centering}p{1.cm}<{\centering}p{1.1cm}<{\centering}p{1.1cm}<{\centering}}  

    \toprule\hline

    \multirow{2}{*}{\bf Setting}&
    \multicolumn{3}{c|}{\bf Visual Features}&
    \multicolumn{3}{c|}{\bf CE Metrics}&
    \multicolumn{4}{c}{\bf NLG Metrcis}\cr
    
    &\bf TXT &\bf GEO  &\bf GLB   &\bf Pre.  &\bf Rec.   &\bf F1  &\bf BL-1   &\bf BL-4  &\bf MTR  &\bf RG-L  \cr

    \midrule

    Baseline &  & & \checkmark  & 0.382 & 0.131 & 0.195 & 44.20 & \textbf{23.65} & 39.94 & 31.53 \cr
    (a) & \checkmark & & & 0.336  & 0.089 & 0.141 & 43.72 & 20.77 & 39.82 & 35.12  \cr
    (b) & \checkmark & \checkmark&  &  \textbf{0.394} & 0.143 & 0.210 & 45.31 & 23.35 & 41.79 & \textbf{37.20} \cr
    (c) & \checkmark & \checkmark & \checkmark & 0.372 & \textbf{0.176} & \textbf{0.239} & \textbf{46.21} & 23.53 &  \textbf{42.94} & 37.01 \cr

    \hline\bottomrule
   
\end{tabular}  
\end{threeparttable} 
\label{tab:ablation} 
\end{table*}

\subsubsection{Global-Local Features Collaboration}
\label{sec:ablation_collaboration}
As shown in Table~\ref{tab:ablation}, we assess the effectiveness of global-local features collaboration by incrementally adding different visual features across settings (a) to (c). Without position and size information, the performance of setting (a) falls significantly below the baseline, underscoring the critical role of geometric information in medical imaging for accurately representing local features. Including geometric features in setting (b) leads to notable improvement, surpassing the baseline in most metrics and demonstrating the value of spatial and structural details. Further incorporating global features in setting (c) boosts performance across most CE and NLG metrics, highlighting the efficacy of global context in capturing inter-regional relationships and improving report coherence. However, we observe that the improvement in recall comes at the expense of precision. This trade-off may result from the influence of abnormality information within global features, which may misrepresent the diagnosis of individual regions and lead to an increase in false positives across regions.

\subsubsection{Region-Report Alignment Training Strategy}
\label{sec:ablation_strategy}
In Table~\ref{tab:rra}, we validate the efficacy of our region-report alignment (RRA) training strategy, which guides the LLM to generate reports based on referring region information for reliable grounding. The results indicate that our model with RRA outperforms the one without it across all CE and NLG metrics, except for ROUGE-L. This improvement demonstrates the effectiveness of the RRA strategy in aligning region-specific features with report generation, ensuring more accurate and clinically relevant outputs. The slightly lower ROUGE-L score can be attributed to its reliance on exact sequence matching, which may overlook the use of synonyms or paraphrased expressions. As a result, ROUGE-L may not fully reflect report quality in cases where the phrasing differs but the semantic content remains consistent. The top METEOR performance supports this, as it better captures semantic similarity by considering synonyms and paraphrasing.

Additionally, the proposed RRA strategy mitigates the precision-recall trade-off in (c). By directing the LLM to reference explicit region information, our model not only ensures that each report is grounded in the correct region but also improves its diagnostic and linguistic quality.

\subsubsection{LLM as the Language Decoder}
\label{sec:ablation_decoder}
To validate the necessity of employing a large-scale LLM as the language decoder for region-level referring and grounding, we conduct a comparative study using GPT-2~\cite{radford2019language} as the decoder. As demonstrated in Table~\ref{tab:decoder}, the LLaMA2-7B model consistently outperforms the GPT-2 model across all metrics, highlighting the advantages of large-scale LLMs in this context. While the GPT-2 decoder performs competently with only global features in report generation~\cite{nicolson2023improving,alfarghaly2021automated}, it struggles to effectively refer to the specific regional features and capture the inter-regional relationships required for region-specific reports. Complex region-level referring and grounding need the powerful in-context learning and long-term referencing abilities of large-scale LLMs, underscoring their importance in generating accurate and contextually rich medical reports tailored to specific anatomical regions.

\setlength{\tabcolsep}{2.5mm}  
\begin{table*}[htp]  
\caption{The effectiveness of region-report alignment (RRA) strategy.}
\centering  
\fontsize{10}{13}\selectfont  
\begin{threeparttable}  
	  
\begin{tabular}{c|p{1.6cm}<{\centering}p{1.2cm}<{\centering}p{1.6cm}<{\centering}|p{1.6cm}<{\centering}p{1.6cm}<{\centering}p{1.8cm}<{\centering}p{2.1cm}<{\centering}}  

    \toprule\hline
    \multirow{2}{*}{\makecell{\bf RRA}}&
    \multicolumn{3}{c|}{\bf CE Metrics}&
    \multicolumn{4}{c}{\bf NLG Metrcis}\cr
    
    &\bf Precision  &\bf Recall   &\bf F1-score  &\bf BLEU-1 &\bf BLEU-4  &\bf METEOR  &\bf ROUGE-L  \cr

    \midrule
    & 0.372 & 0.176 & 0.239 & 46.21 & 23.53 & 42.94 & \textbf{37.01} \cr
    \checkmark & \textbf{0.423} & \textbf{0.181} & \textbf{0.253} & \textbf{47.25} & \textbf{24.87} & \textbf{44.07} & 36.65 \cr

    \hline\bottomrule
   
\end{tabular}  
\end{threeparttable} 
\label{tab:rra} 
\end{table*}

\setlength{\tabcolsep}{2.5mm}  
\begin{table*}[t]  
\caption{The effectiveness of large-scale LLM as the language decoder.}
\centering  
\fontsize{10}{13}\selectfont  
\begin{threeparttable}  
	  
\begin{tabular}{c|p{1.6cm}<{\centering}p{1.2cm}<{\centering}p{1.6cm}<{\centering}|p{1.6cm}<{\centering}p{1.6cm}<{\centering}p{1.8cm}<{\centering}p{2.1cm}<{\centering}}  

    \toprule\hline
    \multirow{2}{*}{\makecell{\bf Language \\ \bf Decoder}}&
    \multicolumn{3}{c|}{\bf CE Metrics}&
    \multicolumn{4}{c}{\bf NLG Metrcis}\cr
    
    &\bf Precision  &\bf Recall   &\bf F1-score  &\bf BLEU-1 &\bf BLEU-4  &\bf METEOR  &\bf ROUGE-L  \cr

    \midrule
   GPT-2~\cite{radford2019language} & 0.357 & 0.169 & 0.229 & 35.47 & 14.72 & 35.76 & 25.18 \cr
    LLaMA2-7B~\cite{touvron2023llama} & \textbf{0.423} & \textbf{0.181} & \textbf{0.253} & \textbf{47.25} & \textbf{24.87} & \textbf{44.07} & \textbf{36.65} \cr

    \hline\bottomrule
   
\end{tabular}  
\end{threeparttable} 
\label{tab:decoder} 
\end{table*}

\subsection{Qualitative Results}
\subsubsection{Report Length Distributions}
\begin{figure}
    \centering
    \includegraphics[width=\linewidth]{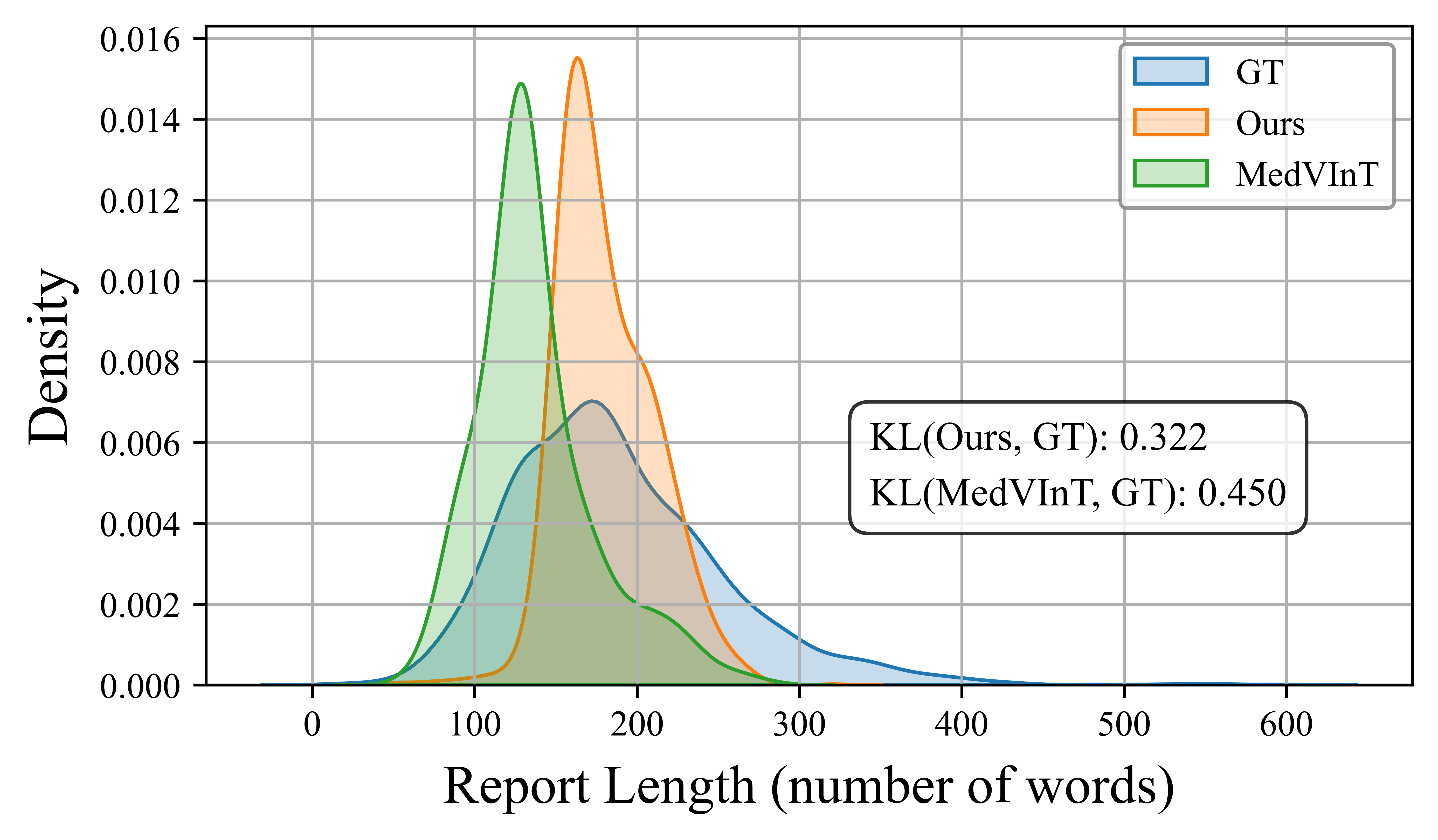}
    \caption{The report length distributions of the ground-truth reports, along with those generated by our proposed method and the SOTA MedVInT~\cite{zhang2023pmc}. The Kullback-Leibler (KL) divergence is utilized to quantify the differences between the distributions of our method and MedVInT relative to the ground-truth reports.}
    \label{fig:length}
\end{figure}

Following \cite{chen2020generating}, we analyze the report length distributions of both generated and ground-truth reports. Fig.~\ref{fig:length} presents the distributions of report lengths for the ground-truth reports alongside those generated by our method and the SOTA MedVInT~\cite{zhang2023pmc}. We leverage Kernel Density Estimation (KDE) to visualize the probability distributions and compute KL divergence to quantify the differences between the distributions of our method and MedVInT relative to the ground-truth reports. The results indicate that our model generates reports with lengths more closely aligned with the ground-truth reports than those generated by MedVInT, as reflected in the lower KL divergence. This suggests that the reports generated by our model are more complete and accurate, whereas MedVInT tends to produce shorter reports, potentially leading to information loss.

\subsubsection{Analysis of Generated Reports}
We present two cases from the RadGenome-ChestCT dataset~\cite{zhang2024radgenome} in Fig.~\ref{fig:case}. Different colors represent the various correctly diagnosed regions described in the reports, while the gray background highlights incorrect diagnoses. Due to the full report length, we focus on the most relevant sections concerning abnormalities in the ground-truth reports.

In the first case, the MedVInT model misdiagnoses multiple regions as normal, missing critical abnormalities. In contrast, our model identifies most abnormalities except the liver lesion. This demonstrates our model’s ability to provide more comprehensive and precise diagnostic information. Notably, it also accurately identifies the location and severity of emphysematous changes in the lungs.

In the second case, our model accurately identifies all abnormalities except for bone degenerative change, whereas MedVInT fails to detect any of them mentioned in the ground-truth report. Although MedVInT points out an abnormality of linear atelectatic changes in the lungs, the diagnosis is likely incorrect as it is not referenced in the ground-truth report. These results demonstrate that our model achieves higher diagnostic accuracy.

\subsubsection{Region-level Reports with Referring and Grounding}
\begin{figure*}
    \centering
    \includegraphics[width=\linewidth]{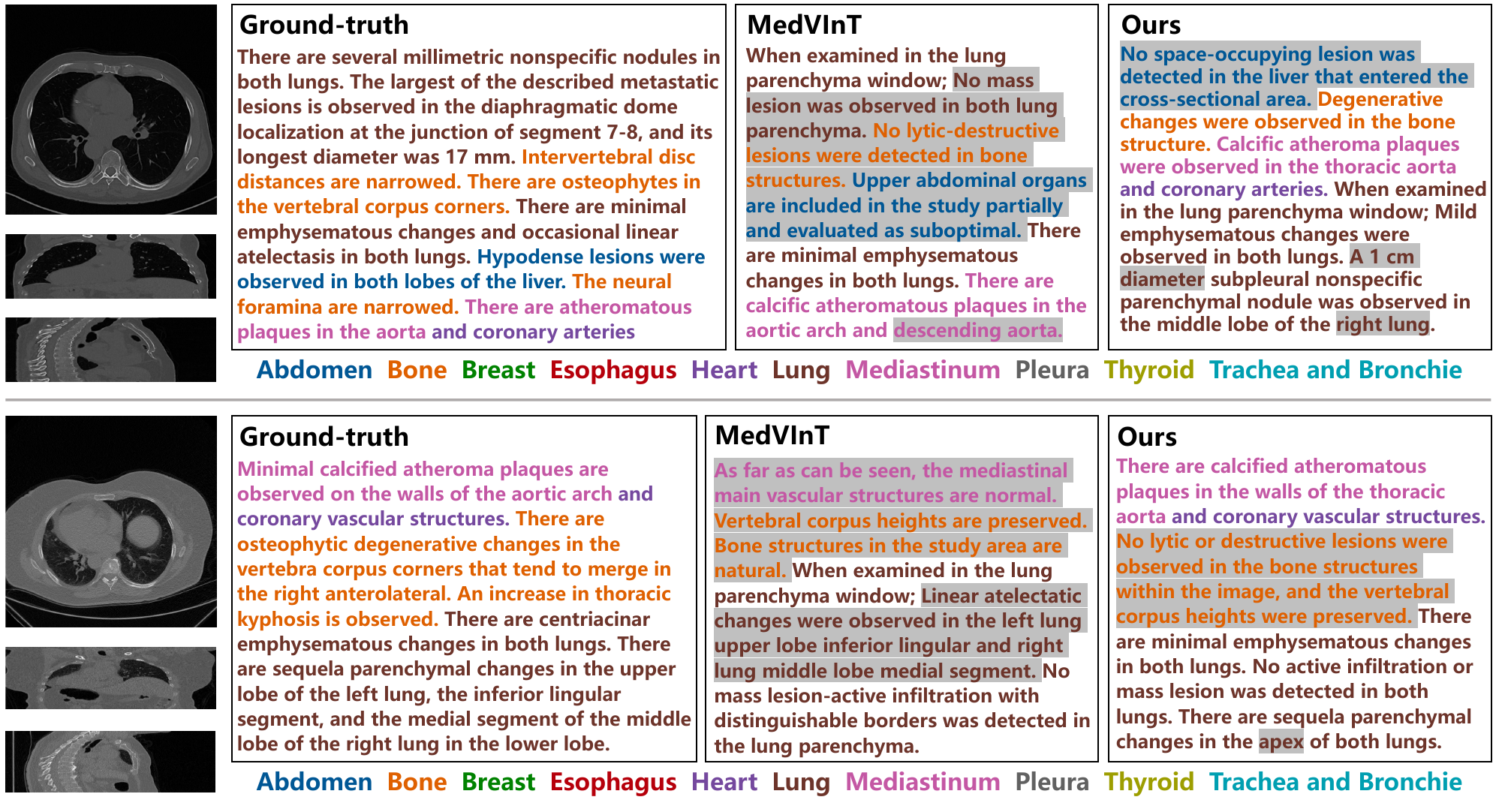}
    \caption{Case studies of our model and the SOTA MedVInT~\cite{zhang2023pmc}. The different colors represent distinct anatomical areas, as shown at the bottom of each example. The \emph{gray} background highlights incorrect diagnoses.}
    \label{fig:case}
\end{figure*}

\begin{figure*}
    \centering
    \includegraphics[width=\linewidth]{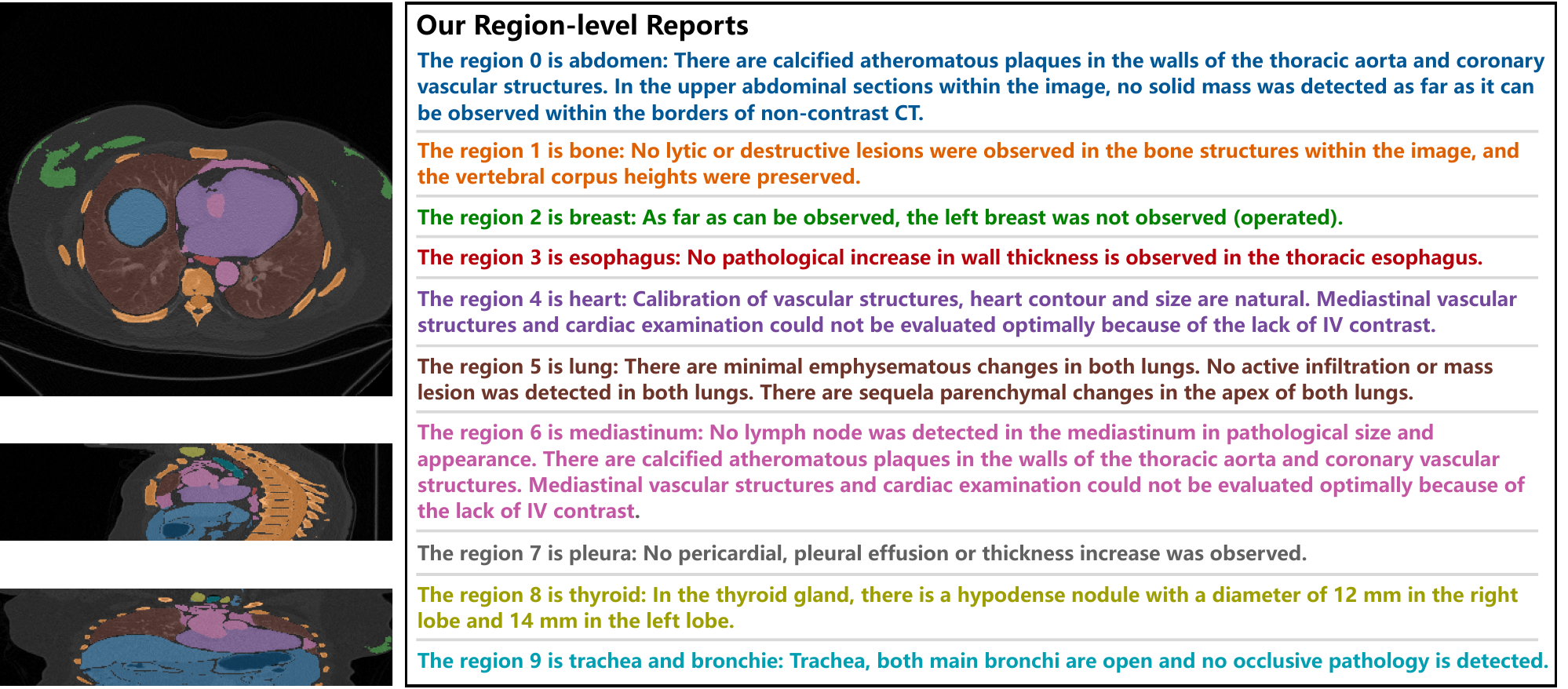}
    \caption{Region-level reports generated by our model. Each regional report refers to a specific region and is grounded in the anatomical area depicted in the left figure. The different colors correspond to distinct anatomical regions.}
    \label{fig:grounding}
\end{figure*}

A key advantage of our model is its ability to generate region-level reports explicitly grounded to specific regions. As illustrated in Fig.~\ref{fig:grounding}, our generated reports are segmented into distinct sections, each corresponding to a particular anatomical region. We use the same color scheme as in Fig.~\ref{fig:case} to represent the region masks and reports associated with different regions. 

Each region-level report begins with the referring region, providing an initial hint about the focus of the LLM. The area recognition result is presented before the report, helping verify the reliability of the generated report. Successfully identifying the referring region indicates that the LLM refers to the correct regional information, enabling the report to be properly grounded in the corresponding region. This enhances the model's interpretability and reliability of the reports, which is valuable for clinical practice. Conversely, if the referenced region is misidentified, the report becomes unreliable regardless of its content and cannot be definitely linked to any region. This strategy provides a straightforward yet effective mechanism for validating reports and offers reliable reference information to assist radiologists in their interpretation.

\subsubsection{Segmentation Results on the Two Datasets}
\begin{figure*}
    \centering
    \includegraphics[width=\linewidth]{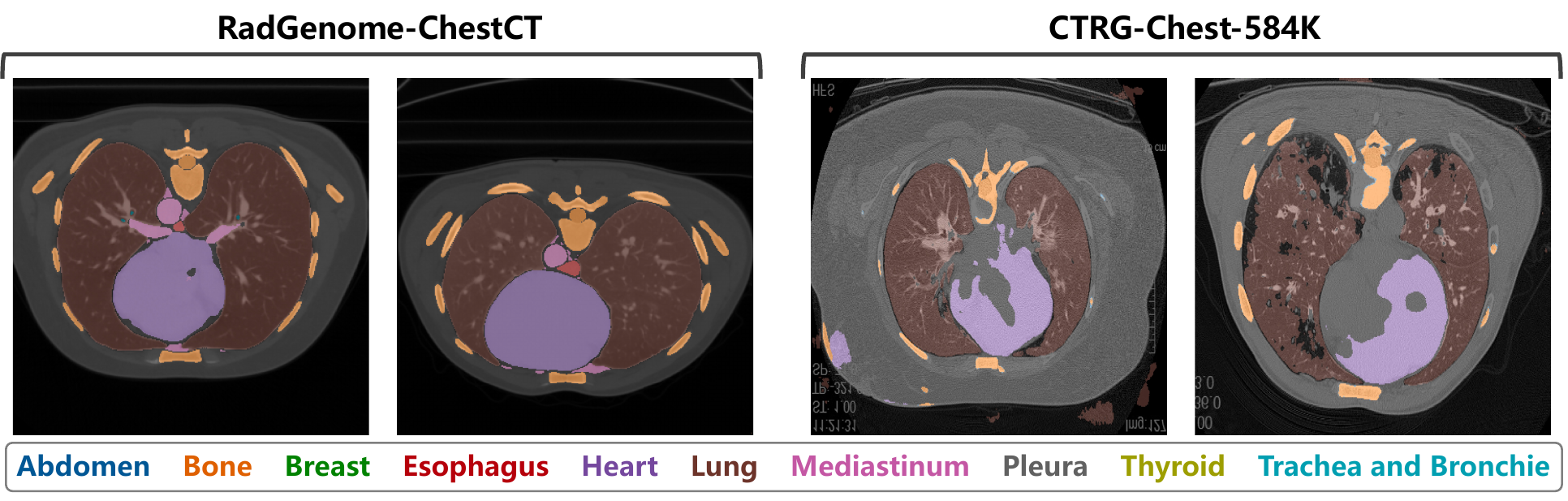}
    \caption{Segmentation results on the RadGenome-ChestCT and CTRG-Chest-584K datasets. The different colors represent distinct anatomical areas, as shown at the bottom.}
    \label{fig:seg_case}
\end{figure*}

We present two segmentation cases from both the RadGenome-ChestCT and CTRG-Chest-584K datasets in Fig.~\ref{fig:seg_case}. Benefiting from the superior performance of SAT~\cite{zhao2023one}, the segmentation results on the RadGenome-ChestCT dataset are highly satisfactory, providing accurate regional information that enables our model to achieve outstanding performance. Regarding the CTRG-Chest-584K dataset, its segmentation results are relatively coarser due to the lower quality of the volumes in this dataset. This may limit the potential of our model, as evidenced by Table~\ref{tab:radgenome}, where its improvement over the second-best method on CTRG-Chest-584K is smaller than that on RadGenome-ChestCT. However, our model still outperforms other methods by effectively leveraging regional information, demonstrating the robustness and effectiveness of our framework.

\section{Limitations and Future Directions}
Despite the encouraging outcomes, several limitations of our model warrant further investigation in the future.

Our framework utilizes the segmentation results to acquire local features. The inaccurate or incomplete masks, caused by segmentation failures, could potentially affect the performance of our model to some extent. Therefore, strengthening the robustness of our method against segmentation errors is valuable for improving the reliability of the generated reports. Given that the relative positions of anatomical regions are generally consistent across patients, incorporating anatomical spatial priors can help refine the anatomical masks. This prior anatomical knowledge enhances the model’s robustness to segmentation failures, thereby ensuring the quality of the extracted local features. Moreover, we plan to explore integrating segmentation model optimization into the training process to further enhance the quality of the anatomical masks, thereby better supporting diagnosis.

Another limitation is the insufficiency of detailed lesion information. The current local features in our framework are restricted to organ-level information, which is inadequate for precise lesion detection. As shown in Fig.~\ref{fig:case}, the model may miss certain lesions or misidentify their locations. To enhance the model's ability to detect and characterize lesions, we plan to explore integrating lesion segmentation or detection into the framework. Incorporating lesion-specific information will enable the model to capture more detailed insights into abnormalities, thereby facilitating the generation of more accurate and clinically relevant reports.

\section{Conclusion}
In this study, we propose the \textbf{Reg2RG} framework for CT report generation. Unlike existing methods relying only on global features, our approach integrates local features with global features, improving the model’s ability to identify detailed lesions in sub-regions, while also enhancing the model’s interpretability and the reliability of the reports. Specifically, we use anatomical masks from a universal segmentation model to capture local features of referring regions. To retain high-resolution local details with low computational cost, a local feature decouple strategy (LFD) is introduced to decouple local features into two parts. Texture features capture fine-grained details of cropped region volumes, while geometric features encode position and size information lost during cropping. The global features are also incorporated to achieve a holistic volume representation. Through the collaboration of global and local features, the model effectively captures the inter-regional relationships while preserving detailed insights within each region. To improve referring and grounding, we propose a training strategy RRA that uses region recognition to guide region-specific report generation. This strategy enhances interpretability and reliability by ensuring reports are grounded in the correct regions. Extensive experiments on two large-scale chest CT datasets demonstrate the superiority of our model over compared methods in both NLG and CE metrics. 

Our work propels CT report generation with a region-guided mechanism, enhancing the trustworthiness of the generated reports. In the future, we plan to extend to additional imaging modalities and anatomical regions while integrating more fine-grained information, aiming to provide a versatile and efficient solution for automated report generation in radiology.

\bibliographystyle{IEEEtran}
\bibliography{main}

\end{document}